# Existence and Global Logarithmic Stability of Impulsive Neural Networks with Time Delay

A. K. Ojha [1], Dushmanta Mallick [2] and C. Mallick [3]

[1] School of Basic Sciences, I.I.T Bhubaneswar
Samantarapuri, Bhubaneswar, Orissa-751013, India

[2] Department of Mathematics, Centurion Institute of Technology
Jatni, Bhubaneswar, Orissa-752050, India

[3] Department of Mathematics, B.O.S.E
Cuttack, Orissa-753007, India

**Abstract**
The stability and convergence of the neural networks are the fundamental characteristics in the Hopfield type networks. Since time delay is ubiquitous in most physical and biological systems, more attention is being made for the delayed neural networks. The inclusion of time delay into a neural model is natural due to the finite transmission time of the interactions. The stability analysis of the neural networks depends on the Lyapunov function and hence it must be constructed for the given system. In this paper we have made an attempt to establish the logarithmic stability of the impulsive delayed neural networks by constructing suitable Lyapunov function.

## 1. Introduction

In recent year's dynamic characteristics of the neural networks has become a focal subject of intensive research studies. Time delay is ubiquitous in most physical and biological systems. In the case of information propagation through a neural network, time delay has been demonstrated to have a substantial influence on the temporal characteristics of the oscillatory behavior of the neural circuits. Jiang [14] proved that time delay can induce multistability, desynchronization, amplitude death and change of pattern in certain dynamical systems. Time delay estimation has diverse application such as in the radar, sonar, seismology, communication system and biomedicine. Shaltaf [23] used constant time delay neural networks to study classification, approximation of nonlinear relation, interpolation and system identification.
The main objective of stability analysis is to find the global exponential stability. It has been established that the sufficient conditions are obtained for the existence and global exponential stability of a unique periodic solution of a class of neural networks with variable and unbounded delays and impulses by using Mawhin's continuous theorem of coincidence degree theory and by constructing Lyapunov function by Yongkun Li [32]. Global exponential stability and periodic solution of Cohen-Grossberg neural networks with continuously distributed delays have been vividly analysed by Li, Y.K [16]. In most situations the delays are variable and unbounded. These types of delay terms suitable for practical neural networks are called unbounded delays. The similar results are also reflected in the studies of [9], [4], [7], [34], [33]. The neural networks can be classified by two categories that are either continuous or discrete but the neural network having not purely continuous or discrete is said to be impulsive neural networks. The characteristic of impulsive neural network is studied by [11], [10], [3], [20], [19].

In the present paper we have made an attempt to study the logarithmic stability of neural networks of periodic solution of a class of neural networks with impulses. The delays used in the neural networks are variable and unbounded. The sufficient conditions are obtained by global logarithmic stability of unique periodic solution of a class of neural networks with variable and by Mawhin's theorem of coincidence degree theory. With reference to this, we determine the unbounded delays, impulses and Lyapunov functions.

Though a lot of works on the stability analysis of delayed neural networks have been made, but the recent survey undertaken by Xu and Lam [24] on sufficient stability of time delay has a great significance in this direction. The delay dependent stability criteria for the linear retarded and neural system with multiple delays have been studied by Park [8] by employing Lyapunov functional approach. More work on the stability analysis on delayed neural system can be found in [13], [31], [30], [15]. Yousefi and Lohmann [2] have studied the instability of neural networks in similarity transformation based model reduction method extend the modification of different reductions methods.

In a recent work Tan and Tan [25] have discussed the exponential stability of neural network where they have considered the variable coefficients and several time varying delays for establishing the uniqueness of the stability of neural networks using periodic activation function with delays. In the high order recurrent neural network Qiu [21] have studied the global stability with time varying delay using bounded activation function.

The organization of the paper is as follows; following the introduction we have used some notations, definitions and results in $2^{nd}$ section. In section 3 the existence of periodic solution are discussed. In section 4 the global exponential stability of periodic solution are presented while in section 5 global logarithmic stability of periodic solution are depicted. Finally in section 6 present the conclusion.

## 2. Preliminaries

The normal neural networks with variable and unbounded time delays and impulses can be defined by integro-differential equation

$$\frac{dx_i(t)}{dt} = -a_i x_i(t) + \sum_{j=1}^{n} \left[ \begin{array}{l} a_i f_j(x_j(t)) \\ + b_{ij} f_j(x_j(t-\tau_{ij}(t))) \\ + c_{ij} \int_{-\infty}^{t} k_{ij}(t-s) \\ f_j(x_j(s)) ds \end{array} \right] + I_i$$

(2.1)

$t > 0, t \neq t_k, i = 1, 2, .......n$

$\Delta x_i(t_k) = I_i(x_i(t_k)) = -\gamma_{ik} x_i(t_k), i = 1, 2, ....... n$

and k=1,2,..n

Where $\Delta x_i(t_k) = x_i(t_k^+) - x_i(t_k^-)$ are the impulses at moments $t_k$ and $0 < t_1 < t_2 < ......$ is strictly increasing sequence.

Such that $\lim_{t \to \infty} t_k = +\infty$

Where $x_i(t)$ is the state of $i^{th}$ neuron; i = 1, 2, …n and n is the number of neurons.
A, B, C are connection matrices and

$$A = \begin{pmatrix} a_{11} & a_{12}..... & a_{1n} \\ a_{21} & a_{22}..... & a_{2n} \\ . \\ . \\ a_{n1} & a_{n2} & a_{nn} \end{pmatrix}, B = \begin{pmatrix} b_{11} & b_{12}..... & b_{1n} \\ b_{21} & b_{22}..... & b_{2n} \\ . \\ . \\ b_{n1} & b_{n2} & b_{nn} \end{pmatrix}$$

$$C = \begin{pmatrix} c_{11} & c_{12}..... & c_{1n} \\ c_{21} & c_{22}..... & c_{2n} \\ . \\ . \\ c_{n1} & c_{n2} & c_{nn} \end{pmatrix}$$

$I = (I_1, I_2,....I_n)^T$ = constant input vector

$$f(x) = \begin{pmatrix} f_1(x) \\ f_2(x) \\ . \\ . \\ f_n(x) \end{pmatrix}$$

f(x) is the activation function of the neurons

$$D = \begin{pmatrix} a_1 & ..... & 0 \\ 0 & a_2... & 0 \\ . \\ . \\ 0 & 0.... & a_n \end{pmatrix}$$

Where $a_i > 0$ and i = 1, 2, ……n
The delays $0 \leq \tau_{ij} \leq \tau$ where i,j=1,2,…n are bounded function.
$k_{ij} : [0,\infty) \to [0,\infty)(i, j = 1,2,..n)$ are piecewise continuous on $[0,\infty)$ and satisfy

**(P1)** $\int_0^{\infty} \log \alpha s k_{ij}(s) ds = p_{ij}(\alpha), (i, j) = 1,2,..n$

Where $p_{ij}(\alpha)$ are continuous function in $[0,\delta), \delta > 0$
and $p_{ij}(0) = 1$

**(P2)** $\int_0^{\infty} k_{ij}(s) ds = 1$

and $\int_0^{\infty} s k_{ij}(s) ds < +\infty$

and i,j = 1, 2,…..n
The condition P1 implies condition P2.
Though $x_i'(t_k)$ does not exist but $x_i'(t_k) \equiv x_i'(t_k^-)$

Though $x'_i(t_k)$ does not exist but $x'_i(t_k) \equiv x'_i(t_k^-)$

The initial condition in (2.1) is of the form $x_i(s) = \phi_i(s)$, $s \leq 0$, $\phi_i$ is bounded and continuous on $(-\infty, 0]$.

Let us assume that:

(A) The delays $0 \leq \tau_{ij} \leq \tau (i, j = 1, 2, \ldots n)$ are bounded

Function with periodic $\omega$ and $a_i > 0, i = 1, 2, \ldots n$

(B) $k_{ij}$ are the piecewise continuous function where i,j=1,2,..n

(C) $f_j \in C(R, R)$, $j = 1, 2, \ldots n$ is Lipschitzian constant and $L_j > 0$, $|f_j(x) - f_j(y)| \leq L_j |x - y|$ for all $x, y \in R$

(D) $M_j > 0$ such that $|f_j(x)| \leq M_j$ for j=1, 2,…n, x∈R where $M_j$ is a positive constant.

(E) There exist a positive integer m such that
$t_{k+m} = t_k + w$
$\gamma_i(k + m) = \gamma_{ik} < 1$ for k = 1, 2,…..n and i = 1, 2, …..n

(F) $\prod_{\sigma \leq t_k \leq t}(1 - \gamma_{ik})$, i = 1, 2, ……n are periodic of $\omega$

Let the impulsive system

$x'(t) = x(t)f(t, x(x - \tau_1(t)), \ldots x(t - \tau_\pi(t))), t \neq t_k, k = 1, 2, \ldots n$

and $\Delta x(x)|_{t=t_k} = I_k(x(t_k^-))$ (2.2)

where $x \in R^n$, $f: R \times R^n \to R^n$ is continuous.
and $f(t + w, x(t - \tau_1(t)), \ldots x(x - \tau_n(t))) = f(t, x(t - \tau_1)(t), \ldots x(t - \tau_n(t)))$

$I_k: R^n \to R^n$, k = 1, 2, …… are continuous

$\tau_i \in ([t_0, \infty), [0, \infty))$ are Lebesgue measurable periodic function of period $\omega$ and $t - \tau_i(t) \to \infty$ as $t \to \infty$, i = 1, 2,…n.

and there exist a positive integer q such that
$t_{k+q} = t_k + \omega$
$I_{k+q}(x) = I_k(x)$ with $t_k \in R$
$t_{k+1} > t_k$, $\lim_{k \to \infty} t_k = \infty$

$\Delta x(t)|_{t=t_k} = x(t_k^+) - x(t_k^-)|$

For $t_k \neq 0$ (k = 1, 2,…..)
$[0, \omega] \cap \{t_k\} = \{t_1, t_2, \ldots, t_q\}$

Here $t_k$ is said to be a point of jumping.
For any $\sigma \geq t_0$
Let $r_\sigma = \min_{1 \leq i \leq n} \inf_{t \geq \sigma} \{t - \tau_i(t)\}$

Let $PC_\sigma$ is the set of functions $\phi: [r_\sigma, \sigma] \to R$ then these are real valued absolute continuous in $[t_k, t_{k+1}] \cap (r_\sigma, \sigma)$.

So $t_k$ placed in $(r_\sigma, \sigma)$ may be discontinuous. So for any $\sigma \geq 0$ and $f \in PC_\sigma$ a function $x \in ([r_\sigma, \infty), R)$ denoted by $x(t, \sigma, \phi)$ is the solution of (2.2) on $(\sigma, \infty)$ and it satisfying the initial condition
$x(t) = \phi(t), \phi(0) > 0, t \in [r_\sigma, \sigma]$ (2.3)
Hence $x(t)$ is absolutely continuous on each interval
$(t_k, t_{k+1}) \subset (r_\sigma, \sigma)$ and for any $t_k \in [\sigma, \infty]$,

k = 1, 2,….. $x(t_k^+)$ and $x(t_k^-)$ exist and $x(t_k^-) = x(t_k)$ and $x(t)$ satisfies (2.2) in $(\sigma, \infty)$ and impulsive point $t_k$ situated in $(\sigma, \infty)$ is discontinuous.

**Definition 2.1.**

The periodic solution $x^*(t)$ of equation (2.1) is said to be globally exponentially stable if there exist constants $\alpha > 0$ and $\beta > 0$ such that
$|x_i(t) - x_i^*(t)| \leq \beta \|\phi - x^*(t)\| e^{(-\alpha t)}|$ for all $t \geq 0$,
where
$\|\phi - x^*(t)\| = \max_{1 \leq i \leq n} \sup_{s \in (-\infty, 0)} |\phi_i(s) - x_i^*(t)|$

To reduce the existence of solution of equation (2.1) for a delay differential equation without impulses

$$\frac{dy_i(t)}{dt} = -a_i y_i(t) + \prod_{0 \leq t_k < t}(1-\gamma_{ik})^{-1} \sum_{j=1}^{n}\left[a_{ij}f_j\left(\prod_{0 \leq t_k < t}(1-\gamma_{jk})y_j(y)\right)\right.$$

$$+ b_{ij}f_j\left(\prod_{0 \leq t_k < t-\tau_{ij}(t)}(1-\gamma_{jk})y_j(t-\tau_{ij}(t))\right) + c_{ij}\int_{-\infty}^{t}k_{ij}(t-s)f_j$$

$$\left.\left(\prod_{0 \leq t_k < s}(1-\gamma_{jk})y_j(s)\right)ds\right] + \prod_{0 \leq t_k < t}(1-\gamma_{ik})^{-1}I_i, t > 0, i = 1, 2, \ldots n$$

(2.4)

with initial condition $y_i(t) = \phi_i(t), t \leq 0$.

**Theorem 2.1.**

Let $\prod_{\sigma \leq t_k \leq t}(1 - \gamma_{ik})$ and i=1,2,….n are periodic function of $\omega$ then

(i) If $y = (y_1, y_2, \ldots y_n)$ is a solution of (2.4) then
$$x = \left(\prod_{0 \leq t_k < t}(1-\gamma_{1k})y_1, \ldots \prod_{0 \leq t_k < t}(1-\gamma_{nk})y_n\right)$$ is a

solution of (2.1)
(ii) If $x = (x_1, \ldots x_n)$ is a solution of (2.1) then
$$y = \left(\prod_{0 \leq t_k < l}(1-y_{1k})^{-1}x_1, \ldots \prod_{0 \leq t_k < t}(1-\gamma_{nk})^{-1}x_n\right)$$ is

solution of (2.4)

**Proof:** As $x_i = \left(\prod_{0 \leq t_k < t}(1-\gamma_{nk})y_i\right)$

is absolutely continuous on the interval $(t_k, t_{k+1})$
and for any $t \neq t_k$, k = 1, 2, …… then

$$x = \left( \prod_{0 \leq t_k < t} (1-\gamma_{1k})y_1, \ldots\ldots\ldots \prod_{0 \leq t_k < t_k} (1-\gamma_{nk})y_n \right)$$

satisfy the system (2.1)
For every $t_k \in \{t_k\}$

$$x_i(t_k^+) = \lim_{t \to t_k^+} \prod_{0 \leq t_j < t} (1-\gamma_{ij})y_i(t) = \prod_{0 \leq t_j < t_k} (1-\gamma_{ij})y_i(t_k)$$

$$x_i(t_k) = \prod_{0 \leq t_j < t} (1-\gamma_{ij})y_i(t_k)$$

for k = 1, 2, …..

$$x_i(t_k^+) = (1-\gamma_{ik})x_i(t_k) \quad (2.5)$$

Which proves (i).
Since $x_i(t)$ is absolutely continuous on each interval $(t_k, t_{k+1})$ and k = 1, 2,….
Then

$$y_i(t_k^+) = \prod_{0 \leq t_j < t_k} (1-\gamma_{ij})^{-1} x_i(t_k^+)$$

$$= \prod_{0 \leq t_j < t_k} (1-\gamma_{ij})^{-1} x_i(t_k) = y_i(t_k)$$

and

$$y_i(t_k^-) = \prod_{0 \leq t_j < t_{k-1}} (1-\gamma_{ij})^{-1} x_i(t_k^-) = y_i(t_k)$$

which implies $y_i(t)$ is continuous on $[o, \infty)$ and also $y_i(t)$ is absolute continuous on $[0, \infty)$ and

$$y = \left( \prod_{0 \leq t_k < t} (1-\gamma_{ik})^{-1} x_1, \ldots\ldots \prod_{0 \leq t_k < t} (1-\gamma_{nk})^{-1} x_n \right) \text{ is a}$$

solution of (2.4).

## 3. Existence of Periodic Solutions

Now we will study the existence of periodic solution by Mawhin's continuation theorem.
Let X, Y are real Banach spaces.
L: Dom L ⊂ X → Y is a linear mapping.
N: X → Y is a continuous mapping.
The mapping L is said to be Fredholm mapping of index zero.
DimKer L = condimImL < ∞ and Im L is closed in Y and there exist continuous projector P : X → X and Q:Y→Y such that Im P = Ker L, Ker = Im(I-Q)
So L |$_{Dom L}$ ∩ $_{ker P}$: (I − P) X → ImL is invertible.
Hence we denote the inverse of mapping by K$_P$. If Ω is an open bounded subset of X then the mapping N is said to be L-compact on $\overline{\Omega}$ if QN ($\overline{\Omega}$) is bounded and K$_P$(I − Q) N: $\overline{\Omega}$ → X is compact.
Since Im Q is isomorphic to Ker L, there exists an isomorphism J: ImQ → Ker L. In order to prove the existence we required the following lemma.

**Lemma 3.1.**
Let Ω ⊂ X be an open bounded set and let N: X → Y be a continuous operator and it is L-compact on $\overline{\Omega}$.
(i) for each λ ∈ (0, 1), x ∈ ∂ Ω ∩ DomL, Lx ≠λNx
(ii) for each x∂Ω ∩ Ker L, QN x ≠ 0 and
deg (JQN, Ω ∩ KerL, 0) ≠ 0
So, Lx = Nx has at least one solution in $\overline{\Omega}$ ∩ DomL

**Theorem 3.1.**
Let (A), (B), (C), (D), (E), (F) hold then the system (2.1) has at least one $\omega$ periodic solution.

**Proof:** Now our aim is to prove the non-impulsive delay differential system (2.4) has a $\omega$ periodic solution. By continuation theorem of coincidence degree theory.
X = Z = {x (t) ∈ C (R, R$^n$): x (t + w) = x (t), t ∈ R, x = (x$_1$, x$_2$, …..x$_n$)$^T$}

with the norm
$$\|x\| = \sum_{k=1}^{n} |x_k|_0$$

$$|x_k|_0 = \sup_{t \in [0,\omega]} |x_k(t)|, k = 1,2,\ldots\ldots n$$

X, Z are Banach spaces

Let L$_x$ = x′ and $P_x = \int_0^w x(t)dt$, x ∈ X

$Q_z = \int_0^\omega z(t)dt$, z ∈ Z and N$_y$=(G$_1$(t), G$_2$(t),….G$_n$(t))$^T$,y∈X
So KerL = {y | y ∈ X, y = h, h ∈ R$^n$}
ImL = {x | x ∈ X, $\int_0^w x(s)ds = 0$ } and
dim KerL = n = codim Im L
It is clear that Im L is closed in Z and L is a fredholm mapping of index zero. So P and Q are continuous projectors satisfying
Im P = Ker L and Im L = Ker Q = Im (I-Q)
Hence K$_p$ : ImL → Ker P domL of L$_P$ has the form

$$K_P(Z) = \int_0^t Z(s)ds - \frac{1}{\omega}\int_0^t \int_0^t Z(x)dsdt$$

Thus,

$$QN_y = \left( \frac{1}{\omega}\int_0^w G_1(t)dt,\ldots\ldots,\frac{1}{\omega}\int_0^w G_n(t)dt \right)^T, y \in X$$

$$\text{and } K_P(I-Q)N_y = \begin{pmatrix} \int_0^t G_1(s)dx \\ \cdot \\ \cdot \\ \int_0^t G_j(s)ds \\ \cdot \\ \cdot \\ \int_0^t G_n(s)ds \end{pmatrix}$$

$$-\begin{pmatrix} \frac{1}{w}\int_0^w \int_0^t G_1(s)dsdt \\ \cdot \\ \cdot \\ \frac{1}{w}\int_0^w \int_0^t G_j(s)dsdt \\ \cdot \\ \cdot \\ \frac{1}{w}\int_0^w \int_0^t G_n(s)dsdt \end{pmatrix} - \begin{pmatrix} \left(\frac{1}{\omega}-\frac{1}{2}\right)\int_0^w G_1(s)ds \\ \cdot \\ \cdot \\ \left(\frac{1}{\omega}-\frac{1}{2}\right)\int_0^w G_j(s)ds \\ \cdot \\ \cdot \\ \left(\frac{1}{\omega}-\frac{1}{2}\right)\int_0^w G_n(s)ds \end{pmatrix}$$

Hence QN and $K_p(I - Q)N$ are continuous and by Arzela Ascoli theorem QN ($\overline{\Omega}$), $K_p(I - Q)$ N ($\overline{\Omega}$) are relatively compact for any open bounded set $\Omega \subset X$.

Therefore N is L-compact on $\overline{\Omega}$ for any open bounded set $\Omega \subset X$.

So for a open bounded subset $\Omega$ for the application of the continuation theorem corresponding to the operator equation $L_x = \lambda N_x$, $\lambda \in (0, 1)$, we have

$$x_i'(t) = \lambda\left\{-a_i x_i(t) + \prod_{0 \leq t_k < t}(1-\lambda_{ik})^{-1}\right.$$

$$\sum_{j=1}^n [a_{ij}f_j(\prod_{0 \leq t_k < t}(1-\gamma_{jk})x_j(t))$$

$$+b_{ij}f_j\left(\prod_{0 \leq t_k < t-\tau_{ij}(t)}(1-\gamma_{jk})x_j(t-\tau_{ij}(t))\right)$$

$$+c_{ij}\int_{-\infty}^t k_{ij}(t-s)f_j\left(\prod_{0 \leq t_k < s}(1-\gamma_{jk})x_j(s)ds\right) + \prod_{0 \leq t_k < t}(1-\gamma_{ik})^{-1}|I_i|\int_0^w \prod_{0 \leq t_k < t}(1-\gamma_{ik})^{-1}dt$$

(3.1)

Where $x \in X$ and $i = 1, 2, \ldots n$

suppose that x (t) = $(x_1(t), t_2(t), \ldots x_n(t))^T \in X$ is a solution of the equation (3.1) for $\lambda \in (0, 1)$
Integrating (3.1) over the interval $[0, \omega]$, we have

$$\int_0^w a_i x_i(t)dt = \int_0^w \prod_{0 \leq t_k < t}(1-\gamma_{ik})^{-1} \sum_{j=1}^n \left[a_{ij}f_i\left(\prod_{0 \leq t_k < t}(1-\gamma_{jk})x_j(t)\right)\right.$$

$$+b_{ij}f_j\left(\prod_{0 \leq t_k < t-T_{ij}(t)}(1-\gamma_{jk})x_j(t-\tau_{ij}(t))\right)$$

$$+c_{ij}\int_{-\infty}^t k_{ij}(t-s)f_j(\prod_{0 \leq t_k < s}(1-\gamma_{jk})x_j(s))ds]dt + I_i \prod_{0 \leq t_k < t}(1-\gamma_{ik})^{-1}dt$$

(3.2)

Let $\bar{t}_i \in [0,\omega] \neq t_k, k=1,2,\ldots m$ such that $x_i(\bar{t}_i) = \sup_{t \in [0,\omega]} x_i(t), i = 1, 2,\ldots n$.

Then by equation (3.2) we have

$$x_i(\bar{t}_i)\omega a_i \geq \int_0^w \prod_{0 \leq t_k < t}(1-\gamma_{ik})^{-1}\sum_{j=1}^n \left[a_{ij}f_j\left(\prod_{0 \leq t_k < t}(1-\gamma_{jk})x_j(t)\right)\right.$$

$$+b_{ij}f_j\left(\prod_{0 \leq t_k < t-\tau_{ij}(t)}(1-\gamma_{jk})x_j(t-\tau_{ij}(t))\right)$$

$$+c_{ij}\int_{-\infty}^t k_{ij}(t-s)f_j\left(\prod_{0 \leq t_k < s}(1-\gamma_{jk})x_j(s)ds\right]dt + I_i\int_0^w \prod_{0 \leq t_k < t}(1-\gamma_{ik})^{-1}dt$$

$$\geq -\int_0^\omega \prod_{0 \leq t_k < t}(1-\gamma_{jk})^{-1}\sum_{j=1}^n \left[\left|a_{ij}f_j\left(\prod_{0 \leq t_k < t}(1-\gamma_{jk})x_j(t)\right)\right|\right.$$

$$+\left|b_{ij}f_j\left(\prod_{0 \leq t_k < t-T_{ij}(t)}(1-y_{jk})x_j(t-\tau_{ij}(t))\right)\right|$$

$$+\left|c_{ij}\int_{-\infty}^t k_{ij}(t-s)f_j\left(\prod_{0 \leq t_k < s}(1-\gamma_{jk})x_j(s)ds\right)\right|\right]dt$$

So
$$x_1(\bar{t_i}) \geq -\frac{N}{a_i}[n(a+b+c)M+I] = -A_i, i = 1, 2,\ldots n \quad (3.3)$$

Where
a = max {|$a_{ij}$|, i, j = 1, 2,....n}
b = max {|$b_{ij}$|, i, j = 1, 2, .....n}
c = max {|$c_{ij}$|, I, j = 1, 2, .....n}
$$M = \max\left\{\sup_{u \in R}|f_i(u)|, i, j = 1, 2,\ldots n\right\}$$
I = max {|$I_i$|, i = 1, 2,……n }

and $N = \max\left\{\prod_{0 \leq t_k < t}(1-\gamma_{ik}^{-1})dt, i = 1, 2,\ldots n\right\}$

Let $(\bar{t_i}) \in [0, w]$, then
$$x_i(\bar{t_i}) = \inf_{t \in [0,\omega]} x_i(t), i = 1, 2,\ldots n$$

Hence
$$x_i(\underline{t_i}) \leq \frac{N}{a_i}[n(a+b+c)M+I]\sigma = A_i, i = 1, 2,\ldots n \quad (3.4)$$

From equation (3.1) we have
$$[x_i(t)\exp(\lambda a_i t)]' = \lambda\left\{\prod_{0 \leq t_k < t}(1-\gamma_{ik})^{-1}\right.$$

$$\sum_{j=1}^{n}[a_{ij}f_j(\prod_{0 \leq t_k < t}(1-\gamma_{jk})x_j(t))$$

$$+ b_{ij}f_j\left(\prod_{0 \leq t_k < t-T_{ij}(t)}(1-\gamma_{jk})x_j(t-\tau_{ij}(t))\right)$$

$$+ c_{ij}\int_{-\infty}^{t}k_{ij}(t-s)f_i\left(\prod_{0 \leq t_k < s}(1-\gamma_{jk})x_j(s)\right)ds\Big]$$

$$+ \prod_{0 \leq t_k < t}(1-\gamma_{ik})^{-1}\Big\}\exp(\lambda a_i t), i = 1, 2,\ldots n$$

$$\int_{0}^{t}\Big|[x_i(t)\exp(\lambda a_i t)]'\Big|dt$$

$$\leq \int_{0}^{w}\left\{\prod_{0 \leq t_k < t}(1-\gamma_{ik})^{-1}\sum_{j=1}^{n}\Bigg[a_{ij}\Bigg|f_i\prod_{0 \leq t_k < t}(1-\gamma_{jk})x_j(t)\Bigg|\right.$$

$$+ b_{ij}\Bigg|f_i\left(\prod_{0 \leq t_k < t-T_{ij}(t)}(1-\gamma_{jk})x_j(t-\tau_{ij}(t))\right)\Bigg|$$

$$+ c_{ij}\int_{-\infty}^{t}k_{ij}(t-s)\Bigg|f_i\left(\prod_{0 \leq t_k < s}(1-\gamma_{jk})x_j(s)\right)\Bigg|ds\Bigg]$$

$$+ \prod_{0 \leq t_k < t}(1-\gamma_{ik})^{-1}|I_i|\Bigg\}\exp(\lambda a_i t)dt, i = 1, 2,\ldots n$$

$$\leq N * [N(a+b+c)M+I]\int_{0}^{\infty}\exp(\lambda a_i t)dt = B_i \quad (3.5)$$

$$N^* = \max\left\{\prod_{0 \leq t_k < t}(1-\gamma_{ik}^{-1}), i = 1, 2,\ldots n\right\}$$

From (3, 4) and (3, 5) we have
$x_i(t) \exp(\lambda a_i t) \leq x_i(\underline{t_i}) \exp(\lambda a_i \underline{t_i}) +$
$\int_{0}^{w}\Big|[x_i(t)\exp(\lambda a_i t)]'\Big|dt$

$\leq A_i \exp(\omega a_i) + B_i = D_i$, i = 1, 2,……n
Since $\exp(\lambda a_i t) \geq 1$ for $\lambda \in (0, 1)$, t $\in$ [0, w] and $x_i(t) \leq D_i$, i = 1, 2, ……….n, then from equation (3.3) and (3.5) we get

$$x_i(t)\exp(\lambda a_i t) \geq x_i(\bar{t_i})\exp(\lambda a_i \bar{t_i}) - \int_{0}^{w}\Big|[x_i(t)\exp(\lambda a_i t)]'\Big|dt$$

$\geq -A_i \exp(wa_i) - B_i = -D_{ij}$, i = 1, 2, ……..n.
Hence $x_i(t) \geq -D_i$, i = 1, 2, ……n

If $A = \sum_{i=1}^{n}D_i + E$

where A is independent of λ and Ω = {x ∈ X : ‖ x(t) ‖ < t}
So Ω satisfies the condition of Lemma 3.1.
When x ∈∂ Ω ∩ Ker L, X = ($X_1$, $X_2$,……$X_n$)$^T$ is a constant vector in $R^n$ with ‖ x ‖ = A. Then

$$QNx = \left(\frac{1}{w}\int_{0}^{\omega}G_1 dt,\ldots,\frac{1}{w}\int_{0}^{\omega}G_n dt\right)^T, x \in X$$

Where
$$G_i = -a_i x_i + \prod_{0 \leq t_k < t}(1-\gamma_{ik})^{-1}\sum_{j=1}^{n}\Bigg[a_{ij}f_j\left(\prod_{0 \leq t_k < t}(1-\gamma_{jk})x_j\right)$$

$$+ b_{ij} f_j \left( \prod_{0 \le t_k < t - T_{ij}(t)} (1 - \gamma_{jk}) x_j \right)$$

$$+ c_{ij} \int_{-\infty}^{4} k_{ij}(t-x) f_j \left( \prod_{0 \le t_k < s} (1 - \gamma_{jk}) x_j \right) ds \Bigg]$$

$$+ \prod_{0 \le t_k < t} (1 - \gamma_{ik})^{-1} I_i, i = 1, 2, \ldots n$$

Let $J : \text{Im} Q \to \text{Ker } L$, $r \to r$
If A is greater than $x^T$ JQN $x < 0$
so for $x \in \partial \Omega \cap \text{Ker } L$, Q N x ≠ 0
Let $\phi(\gamma : x) = -\gamma x + (1 - \gamma)$ JQN x
then for $x \in \partial \Omega \cap \text{Ker } L$, $x^T \phi(\gamma, x) < 0$
So deg{JQN, $\Omega \cap$Ker L, 0} = deg{−x, $\Omega \cap$ Ker L, 0} ≠ 0
for each $x \in \partial \Omega \cap \text{Ker } L$, QN x ≠ 0 and
deg (JQN, $\Omega \cap$ Ker L, 0) ≠ 0
Hence equation (2.4) has at least one $\omega$-periodic solution and system (2.1) has at least one $\omega$ periodic solution.

Before going to study the stability condition of neural networks with time delay we have stating some of the important results due to Mawhin [12] on coincidence degree for perturbations of Fredholm mapping.

**Proposition 3.1.**
Let X, Z be a vector space, dom L a vector subspace of X and L=dom L $\subset$ X $\to$ Z
a linear mapping. Its kernel $L^{-1}(0)$ will be denoted by Ker L and its range L (Dom L) by Im L.
Let P:X$\to$ X,Q:Z$\to$Z be algebraic projectors such that the following sequence is exact :
X $\to$ dom L $\to$ Z $\to$ Z
which mean that ImP=KerL and ImL=KerQ
If we define L$_P$:domL$\cap$KerP$\to$ ImL
as the restriction L|domL$\cap$KerP of L to domL $\cap$ Ker P, then it is clear that L$_P$ is an algebraic isomorphism we shall define K$_P$ : ImL $\to$ domL by $K_P = L_P^{-1}$
Clearly, $K_P$ is one-to-one and $PK_P = 0$
Therefore, on Im L,
LK$_P$=L(I − P)K$_P$=L$_P$(I−P)K$_P$=L$_P$K$_P$=1 and, on dom L,
K$_P$L=K$_P$L(I−P)=K$_P$L$_P$(I−P)=1−P

**Preposition. 3.2.**
Let Coker L= $Z/\text{Im} L$ be the quotient space of Z under the equivalence relation z ~ z' $\Leftrightarrow$ z − z' $\in$ ImL
Thus, Coker L={z + ImL : z $\in$ Z} and we shall denote by $\Pi$ : Z $\to$ Coker L, z $\to$ z + ImL the canonical surjection.

**Proposition. 3.3.**
If there exist a one-to-one linear mapping A:CokerL$\to$ Ker L then the equation L$_x$ = y, y $\in$ Z is equivalent to equation (I − P) x = ($\wedge\Pi$+ K$_{P, Q}$) y where K$_{P,Q}$ : Z $\to$ X is defined by K$_{P, Q}$ = K$_P$(I − Q).
More on this work refer Mawhin [12].

## 4. Global exponential stability of the periodic solution

Suppose x*(t) = (x*$_1$(t), x*$_2$(t),……….x*$_n$(t))$^T$ is a Periodic of system (2.1). In this section some Lyapunov functions are defined to study the exponentially stability of this periodic solution.

**Theorem 4.1.**
Let A – F hold and
(i)There exist n positive constant $\xi > 0$, i = 1, 2,…..n such that

$$-\xi_i a_i + \sum_{j=1}^{n} \xi_j \left( |a_{ij}| + |b_{ij}| + |c_{ij}| \right) L_j < 0, i = 1, 2, \ldots n$$

(4.1)

(ii)The impulses operator I$_i$(x$_i$(t)), i = 1, 2,….n satisfy
I$_i$(x$_i$(t$_k$)) = $-\gamma_{ik}$ (x$_i$(t$_k$) $x_i^*(t)$), 0 < y$_{ik}$<1,i= 1, 2,…..n, k$\in$ Z$^+$

**Proof :** We know that system (2.1) has an $\omega$ periodic solution x*(t) = $\left( x_1^*(t), x_2^*(t), \ldots x_n^*(t) \right)^T$
Let x(t)=(x$_1$(t),x$_2$(t),……x$_n$(t))$^T$ is arbitrary solution of (2.1).If a is $\alpha$ constant satisfying $\delta \ge \alpha > 0$ such that for i=1, 2,…..n then

$$\xi_i(-a_i + a) + \sum_{j=1}^{n} \xi_i \left( |a_{ij}| + e^{aT} |b_{ij}| + |c_{ij}| p_{ij}(\alpha) \right) k_j < 0$$

(4.2)

Let $y(t) = x(t) - x^*(t)$ then the equation (2.1) becomes

$$\frac{dy_i(t)}{dt} = -a_i y_j(t) + \sum_{j=1}^{n} [a_{ij} g_j(y_j(t)) + b_{ij} g_j(y_j(t - \tau_{ij}(t)))$$

$$+ c_{ij} \int_{-\infty}^{t} k_{ij}(t-s) g_j(y_j(s)) \Big] ds \quad (4.3)$$

also $\Delta y_i(t_k) = -\gamma_{ik} y_i(t_k)$, i = 1, 2,……n
and |y$_i$(t$_k$ + 0)| = |1 − $\gamma_{ik}$|| y$_i$(t$_k$) |≤| y$_i$(t$_k$)|
where g$_j$(y$_j$(t)) = f$_j$(x$_j$(t)) − $f_j(x_j^*(t))$, j = 1, 2,………n
By assumption (C), we know that 0 ≤ |g$_i$(y$_i$)| ≤ L$_i$ |y$_i$|, i=1, 2,…..n
The initial condition of (4.3) is $\psi(s) = \phi(s) - x^*(t)$
Let the Lyapunov function V = (V$_1$, V$_2$, ……V$_n$)$^T$ defined by V$_i$ = e$^{\alpha t}$ |y$_i$(t)|, i=1,2,….n then from equation (4.3), we get

$$\frac{d^+ V_i(t)}{dt} = e^{\alpha t} \text{sgn} y_i \{-a_i y_i(t) + \sum_{j=1}^{n} [a_{ij} g_j(y_j(t)) + b_{ij} g_j(y_j(t - \tau_{ij}(t)))$$

$$+ c_{ij} \int_{-\infty}^{t} k_{ij}(t-s) g_j(y_j(s)) ds \} + \alpha e^{\alpha t} |y_i(t)|$$

$$\leq e^{\alpha t}\left\{(-a_i+\alpha)|y_i(t)|+\sum_{j=1}^{n}L_j\left[|a_{ij}||y_j(t)|+\right.\right.$$

$$\left.\left.|b_{ij}||y_j(t-\tau_{ij}(t))|+|c_{ij}|\int_{-\infty}^{t}k_{ij}(t-s)|y_j(s)|ds\right]\right\}$$

$$\leq (-a_i+\alpha)e^{\alpha t}|y_i(t)|+\sum_{j=1}^{n}L_j\left[|a_{ij}|\left|e^{\alpha t}y_j(t)\right|\right.$$

$$+e^{\alpha T_{ij}(t)}|b_{ij}|\left|e^{\alpha(t-T_{ij}(t))}y_j(t-\tau_{ij}(t))\right|$$

$$\left.+|c_{ij}|\int_{-\infty}^{t}k_{ij}(t-s)e^{\alpha(t-s)}\left|e^{\alpha T}y_j(s)\right|ds\right] \quad (4.4)$$

$$\leq (-a_i+\alpha)V_i(t)+\sum_{j=1}^{n}L_j\left[|a_{ij}|V_j(t)+e^{\alpha T}|b_{ij}|\right.$$

$$V_j(t-\tau_{ij}(t))+|c_{ij}|\int_{-\infty}^{t}k_{ij}(t-s)e^{\alpha(t-s)}V_j(s)ds \quad (4.5)$$

For $t > 0$ and $t \neq t_k$.
Defining there curve $\rho=\{w(l):w_i=\xi_i l, l > 0, i = 1, 2,\ldots n\}$
and the set $\Omega(w) = \{u : 0 \leq u \leq w, w \in \rho\}$
$S_i(w)=\{u\in\Omega(w):u_i = w_i, 0 \leq u \leq w\}$ then $l > l$, $\Omega(w(l))$
So the equation (4.3) is exponentially stable.
If there exist a constant $\beta > 0$ and $\alpha > 0$, such that
$\|y(t)\| \leq \beta e^{-\alpha t}\|\psi\|$ for all $t \geq 0$ and
$\xi_{max} = \max_{1\leq i\leq n}\{\xi_i\}$, $\xi_{min} = \min_{1\leq i\leq n}\{\xi_i\}$

$$l_0 = \frac{(1+\delta)\|\psi\|}{\xi_{min}} \text{ where } \sigma > 0 \text{ is a constant.}$$

Then $\{|V| : |V| = e^{\alpha s}|\psi(s)|, -\infty \leq s \leq 0\} \subset \Omega(w_0(l_0))$
and $|V_i(s)| = e^{\alpha s}|\psi_i(s)| < \xi_i l_0, -\infty \leq s \leq 0, i = 1, 2,\ldots n$
so $|V_i(t)| < \xi_i l_0$ for $t \in [0, -\infty], i = 1, 2, \ldots n$
and if it is not true then there exist some i and $t_1$ ($t_1 > 0$)
such that $|V_i(t_1)| = \xi_i l_0$
$D^+|V_i(t_1)| \geq 0$ and $|V_j(t)| \leq \xi_i l_0$ for $-\infty < t \leq t_1$,
$j = 1, 2,\ldots n$ so from equation (4.4) we get

$$D^+V_i(t_1) \leq \left[\xi_i(-a_i+\alpha)+\sum_{j=1}^{n}K_j\left(|a_{ij}|\right.\right.$$

$$\left.\left.+e^{\alpha\tau}|b_{ij}|+|c_{ij}|p_{ij}(\alpha)\xi_j\right]l_0 < 0 \quad (4.6)$$

For $t>0$ and $t \neq t_k$
gives a contradiction
So $|v_i(t)| < \xi_i l_0$ for $t \geq 0, t \neq t_k$

and $v_i(t_k+0) = e^{\alpha t}|y_i(t_k+0)| \leq e^{\alpha t}|y_i(t_k)| = v_i(t_k)$
for $k \in z^+$

and $|y_i(t)| < \xi_i l_0 e^{-\alpha t} \leq (1+\delta)\|\psi\|\frac{\xi_{max}}{\xi_{min}}e^{-\alpha t}$, i=1,2..n

for $t \geq 0$

where $\beta = (1 + \delta) \xi_{max} \xi_{min}$
Hence the periodic solution of system (4.3) is globally exponentially stable.

## 5. Global Logarithmic stability of the periodic solution

**Theorem 5.1.** If the theorem (4.1) holds then the system (4.3) is logarithmically stable.
**Proof: By** assumption (C), we know that $0 \leq |g_i(y_i)| \leq L_i |y_i|$, $i = 1, 2,\ldots n$.
The initial condition of (4.3) is $\psi(s) = \phi(s) - x^*(t)$
Let the Lyapunov function $V=(V_1,V_2\ldots V_n)^T$ defined by
$V_i = \text{Log}\alpha t |y_i(t)|$, $I = 1, 2,\ldots n$ then from equation (4.3), we get

$$\frac{d^+V_i(t)}{dt} = \log\alpha t \, \text{sgn}\, y_i\left\{-a_i y_i(t)+\sum_{j=1}^{n}\left[a_{ij}g_j(y_j(t))+\right.\right.$$

$$b_{ij}g_j(y_j(t-\tau_{ij}(t)))$$

$$\left.\left.+c_{ij}\int_{-\infty}^{t}k_{ij}(t-s)g_j(y_j(s))ds\right]\right\}+\alpha e^{\alpha t}|y_i(t)|$$

$$\leq Log\alpha t\left\{(-a_i+\alpha)|y_i(t)|+\sum_{j=1}^{n}L_j\left[|a_{ij}||y_j(t)|\right.\right.$$

$$\left.\left.+|b_{ij}||y_i(t-\tau_{ij}(t))|+|c_{ij}|\int_{-\infty}^{t}k_{ij}(t-s)|y_j(s)|ds\right]\right\}$$

$$\leq (-a_i+\alpha)|Log\alpha t y_i(t)|+\sum_{j=1}^{n}L_j\left[|a_{ij}|\left|e^{\alpha t}y_j(t)\right|\right.$$

$$+Log\alpha\tau_{ij}(t)|b_{ij}|Log\alpha(t-\tau_{ij})(t)y_i(t-\tau_{ij}(t))|$$

$$\left.+|c_{ij}|\int_{-\infty}^{t}k_{ij}(t-s)Log\alpha(t-s)Log\alpha\tau y_j(s)ds\right] \quad (5.1)$$

$$\leq (-a_i+\alpha)V_i(t)+\sum_{j=1}^{n}L_j\left[|a_{ij}|V_j(t)+Log\alpha\tau|b_{ij}|V_j(t-\tau_{ij}(t))\right.$$

$$+|c_{ij}|\int_{-\infty}^{t}k_{ij}(t-s)e^{\alpha(t-s)}V_j(s)ds| \quad (5.2)$$

for $t > 0$ and $t \neq t_k$.
Defining the curve $\rho=\{w(l):w_i=\xi_i l, l>0, i = 1, 2,\ldots\ldots n\}$
and the set $\Omega(w) = \{u : 0 \leq u \leq w, w\in\rho\}$
$S_i(w)=\{u\in\Omega(w):u_i=w_i, 0\leq u\leq w\}$ then
$l > \tilde{l}, \Omega(w(l))\subset\Omega(w(\tilde{l}))$.
So the equation (4.3) is Logarithmically stable.
If there exist a constant $\beta>0$ and $\alpha>0$, such that
$\|y(t)\| \leq \beta e^{-\alpha t}\|\psi\|$ for all $t \geq 0$
and $\xi_{max} = \max_{1\leq i\leq n}\{\xi_i\}$ and $\xi_{min} = \min_{1\leq i\leq n}\{\xi_i\}$

$$l_0 = \frac{(1+\delta)\|\psi\|}{\xi_{min}}$$

Where δ >0 is a constant.

then $\{|V|:|V| = \log\alpha s |\psi(s)|, -\infty \leq s \leq 0\} \subset \Omega(w_0(l_0))$

then $|V_i(s)| = \log\alpha s |\psi_t(s)| < \xi_i l_0, -\infty \leq s \leq 0, i = 1,2,........n$

so $|V_i(t)| < \xi_i l_0$ for $t \notin [0, -\infty], ,i = 1,2,........n$

and if it is not true then there exist some i and $t_1$ ($t_1 > 0$) such that $|V_i(t_1)| = \xi_i l_0$

$D^+|V_i(t_1)| \geq 0$ and $|V_i(t)|\xi_i l_0$ for $-\infty < t \leq t_1, j = 1,2,....n$

From equation (4.4) we get

$$D^+V_i(t_1) \leq \left[\xi_i(-a_i + \alpha) + \sum_{j=1}^{n} K_j(|a_{ij}| + e^{\alpha\tau}|b_{ij}|)\right]$$

$+ |c_{ij}| p_{ij}(\alpha)\xi_i]l_o <0$ \hspace{2em} (5.3)

for t > 0 and t≠ $t_K$ gives a contradiction

So $|V_i(t)| < \xi i l_o$ for $t \geq 0, t \neq t_k$

and $V_i(t_k+0) = \text{lag}\alpha t |y_i(t_k+0)| e^{\alpha t} |y_i(t_k)| = V_i(t_k)$ for $\kappa \notin Z^+$

and $|y_i(t)| < \xi_i l_0 \text{Log}(-\alpha t) \leq (1+\delta)\|\Psi\| \frac{\xi_{max}}{\xi_{min}}$ Log $(-\alpha t)$,

i=1, 2,.......n for t≥ 0 where β = (1 +δ) $\xi_{max} \xi_{min}$

Hence the periodic solution of system (4.3) is globally exponentially stable.

The following example explain the existence and stability of neural network.

**Example. 1.**

Let us consider the Hopfield neural Network with time delay

$$\frac{dv_i(t)}{dt} = -2.5 \log 3ty(t) + 2g(y_1(t))+2.4g(y_1(t)-1)$$

$+.5g(y_2(t)) +1.5g(y_2(t)-1)+3 \int_{-\infty}^{t}(t-1) g(y(s))ds +3$

where g(y(t))= log(3y(t) + 1), I = 3 for i = 1

Using the above theorem through direct calculation we have

$$\frac{dv_i}{dt} \leq (3 + 2.5)+\{1.5+ 2.4 \log (3y(t)+1)\}+|P_{ij}(3)|l_0 \leq \xi\, l_0$$

for $|_{P_{ij}}(3)| = |\int_0^{\infty} \log (\log 3t +1)dt| = 3$ for y(t) = 3

for the finite value of t

$$\frac{dv_i}{dt} \leq 6.5+2.4 \, \log (3t+1)+3l_0$$

which shows the network global logarithmic stable.

## 6. Conclusion

In this paper we have generalized the work of Youngkun Li[32] to study the existence and global exponential stability of Periodic solution of class of neural networks and also studied the logarithmic stability of neural networks. The sufficient condition generates the existence, unique periodic solution and global logarithmic stability of an equilibrium point by using Mawhin's continuation theorem.

**References**

[1]Arik.S,Tavanoglu.V(1998):Equilibrium analysis of delayed CNNs.IEEE Trans Circuits Syst-I 1998,45:168-71.

[2]Amirhossein Youse, Boris Lohman (2009): A note on stability in model reduction. International Journal of system Science 2009, Vol 40, Issue 3:297-307.

[3]Akca.H,Alassar.R,Covachev.V,Covachev.Z,AlZahrain. E(2004): Continuous time additive Hopfield type neural networks with impulses.J.Math.Anal.Appl.2004, 290:436-51.

[4]Chen.Y (2002): Global stability of neural networks with distributed delays. Neural Networks 2002,15:867-71.

[5]Chunguang.L, Guanrong, Chen, Xiaofeng.l and Juebang.Yu (2004): Hopf Bifeercation and chaos in a single Inertial Neuron model with time delay.Europen Physical Journal B, Vol 44,337-343.

[6]Forti.M and Tesi.A (1995): New conditions for global stability of neural networks with application to linear and quadratic programming problems. IEEE Trans. Circuit Syst.1,42 (7).354-66.

[7]Feng.C, Plamondon.R (2001): On the stability analysis of delayed neural networks systems. Neural Networks.14, 1181-8.

[8]Gwang Seok Park,Ho Lim Choi,Jong Tae Lim(2008):On stability of linear time delay systems with multiple delays. International Journal of System Science.Vol 39, Issue 8,839-852.

[9]Gapalsamy.K, He.X (1994): Stability in asymmetric Hopfield nets with transmission delays.Physica D 1994, 76:344-58.

[10]Guan.Z.H, Chen.G and James.L (2000): On impulsive auto-associative neural networks. Neural Networks.13:63-9.

[11]Guan.Z, Chen.G and Qin.Y (2000): On equilibria, stability and instability of Hopfield neural networks.IEEE Trans Neural Networks 2:534-40.

[12]Gaines.RE, Mawhin.JL (1977): Coincidence degree and nonlinear differential equations.Berlin.Springer-Verlag.

[13]Jun Yoneyama, Takuo Tsuchiya (2008): New delay dependent conditions on robust stability and stabilisation for discrete time systems with time delay. International Journal of System Science.Vol 39, Issue 10, 1033-1040.

[14]Jiang.Y (2000): Physics Letter. A, 267-342.

[15]Liu.Y, Wang.Z and Liu.X (2009): Asymptotic stability for neural networks with mixed time delays: The discrete time case. Neural Networks.Vol 22, 67-74.

[16] Li.Yk, Liu.P (2004): Existence and stability of periodic solution for Cohen-Grossberg Neural Networks with multiple delays. Chaos solutions and fractals, 2004, 20:459-66.

[17] Li.Yk, Liu.P (2004): Existence and stability of periodic solution for BAM Neural Networks with delays. Math Comput Model, 2004, 40:757-70.

[18] Li.YK, Liu.CC, L.F (2005): Global exponential stability of periodic solution for shunting inhibitory CNNs with delays. Phys Lett A 2005, 333, 46-54.

[19] LiYK, Lu.LH (2004): Global exponential stability of periodic soltion of Hopfield type neural networks with impulses. Phys Lett.A, 333:51-61.

[20] LiYK (2005): Global exponential stability of BAM neural networks with delays and impulses. Chaos, Solitions and Fractals.24 (1):279-85.

[21] Qiu.F, Cui.B, Wu.W (2009): Global exponential stability of high order recurrent neural network with time varying delays. Applied Mathematical Modeling. Vol 33,198-210.

[22] Qiu.J (2007): Exponential stability of impulsive neural networks with time-varying delays and reaction diffusion terms. Neuro Computing. Vol 70, 1102-1108.

[23] Shaltaf.S (2004): Neural Network based time delay estimation. EURASIP Journal on applied signal processing.3, 378-385.

[24] Shengyuan Xu, James Lam (2008): A survey of linear matrix techniques in stability analysis of delay systems. International Journal of System Science. Vol 39, Issue 12, 1095-1113.

[25] Tan.M, Tan.Y (2009): Global exponential stability of periodic solution of neural network with variable coefficient and time varying delays. Applied Mathematical Modelling. Vol 33,373-385.

[26] Sun.J, Zhang.Y (2005): Stability of impulsive neural networks with time delays. Physics Letters A, Vol 348, 44-50.

[27] Tavanoglu.V, Arik.S (2000): On the global asymptotic stability of delayed cellular neural networks. IEEE Trans Circuits Syst-I.2000, 47:571-4.

[28] Wang.X, Guo.Q, Xu.D (2009): Exponential P-stability of impulsive stability stochastic Cohen Grossberg neural networks with mixed delays. Mathemaics and Computers in Simulation. Vol 79, 1698-1710.

[29] Xu.S, Lam.J (2006): A new approach to exponential stability analysis of neural networks time varying delays. Neural Networks.19, 76-83.

[30] Xiao He, Zidong Wang, Donghua Zhou (2008): Neural Networked fault detection with random communication delays and packet losses. International Journal of System Science. Vol 39, Issue 11, 1045-1054.

[31] Yuanqing Xia, Jiqing Qiu, Jinhui Zhang, Zhifeng Gao (2008): Delay dependent robust H1 control for uncertain stochastic time delay system. International Journal of System Science. Vol 39, Issue 12, 1139-1152.

[32] Yongkun Li, Wenya Xing, Linghong Lu (2006): Existence and global exponential stability of periodic solution of a class of neural networks with impulses. Neural Network 27.437-445.

[33] Zhang.J, Jin.X (2000): Global stability analysis in delayed Hopfield neural networks models. Neural Networks.13, 745-53.

[34] Zhang.J (2003): Globally exponential stability of neural networks with variable delays. IEEE Trans on Circuit Systems I, 50,288-291.

[35] Zhou.E.E and Erent.R.F (1995): A parallel ordering algorithm for efficient one side Jacobi SVD computations. Proc. of 3rd Euromacro workshop on parallel and distributed processing,San Remo,Italy.

**Dr.A.K.Ojha**: Dr A.K.Ojha received a Ph.D (Mathematics) from Utkal University in 1997.Currently he is an Asst. Prof. in Mathematics at I.I.T Bhubaneswar, India. He is performing research in Neural Network, Genetical Algorithm, Geometric Programming and Particle Swarm Optimization. He is served more than more than 27 years in different Govt. colleges in the state of Orissa. He is published 22 research paper in different journals and 7 books for degree students such as: Fortran 77 Programming, A text book of Modern Algebra, Fundamentals of Numerical Analysis etc.

**Dushmanta Mallick**: Mr.D.Mallick received a M.Phil (Mathematics) from Sambalpur University in 2002.Currently he is an Asst. Prof. in Mathematics at Centurion Institute of Technology, Bhubaneswar, India. He is performing research in Neural Network and Optimization Theory. He is served more than more than 6 years in different Colleges in the state of Orissa. He is published 8 research paper in different journals.

**Dr.C.Mallick**: Dr.C.Mallick received a Ph.D (Mathematics) from Utkal University in 2008.Currently he is an Lecturer in Mathematics at B.OS.E Cuttack, India. He is performing research in Neural Network. He is published 3 books for degree students such as: A Text Book of Engineering Mathematics, Interactive Engineering Mathematics etc.